\pdfoutput=1

\documentclass[11pt]{article}

\usepackage[]{acl}
\usepackage{multirow}
\usepackage{tabularx}
\usepackage{colortbl}
\usepackage{times}

\usepackage{latexsym}
\usepackage{booktabs}

\usepackage[T1]{fontenc}

\usepackage[utf8]{inputenc}
\usepackage{CJKutf8}
\usepackage{microtype}
\usepackage{xurl}
\usepackage{hyperref}
\usepackage{breakurl}

\usepackage{inconsolata}

\usepackage{graphicx}
\PassOptionsToPackage{table}{xcolor}\usepackage{xcolor}
\usepackage{xspace}
\newcommand{\ours}{{EasyEdit2}\xspace}

\title{\raisebox{-10.0pt}{\includegraphics[scale=0.15]{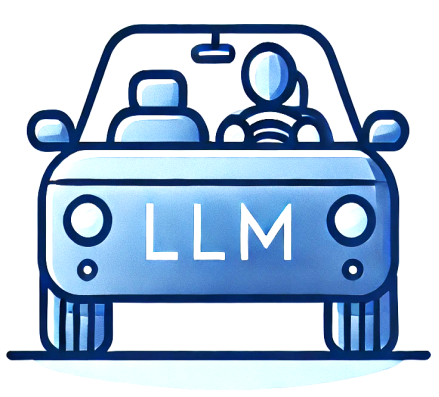}} EasyEdit2: An Easy-to-use Steering Framework for Editing Large Language Models}

\author{First Author \\
  Affiliation / Address line 1 \\
  Affiliation / Address line 2 \\
  Affiliation / Address line 3 \\
  \texttt{email@domain} \\\And
  Second Author \\
  Affiliation / Address line 1 \\
  Affiliation / Address line 2 \\
  Affiliation / Address line 3 \\
  \texttt{email@domain} \\}

\author{
Ziwen Xu\textsuperscript{1},
Shuxun Wang\textsuperscript{1},
Kewei Xu\textsuperscript{1},
Haoming Xu\textsuperscript{1},
Mengru Wang\textsuperscript{1},
{\bf Xinle Deng}\textsuperscript{1},\\
{\bf Yunzhi Yao}\textsuperscript{1},
{\bf Guozhou Zheng}\textsuperscript{2},
{\bf Huajun Chen}\textsuperscript{1},
{\bf Ningyu Zhang}\textsuperscript{1}\thanks{~Corresponding author.}\\
\textsuperscript{1}Zhejiang University
\\
\textsuperscript{2}Ocean Research Center of  Zhoushan, Zhejiang University\\
  \texttt{\{ziwen.xu,zhangningyu\}@zju.edu.cn}\\
  \raisebox{-1.2pt}{\includegraphics[scale=0.03]{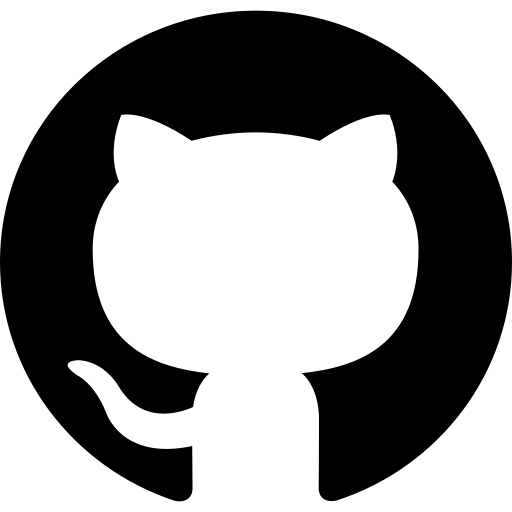}}\,\url{https://zjunlp.github.io/project/EasyEdit2}
}

\begin{document}

\maketitle
\begin{abstract}
In this paper, we introduce EasyEdit2, a framework designed to enable plug-and-play adjustability for controlling Large Language Model (LLM) behaviors. EasyEdit2 supports a wide range of test-time interventions, including safety, sentiment, personality, reasoning patterns, factuality, and language features. Unlike its predecessor, EasyEdit2 features a new architecture specifically designed for seamless model steering. It comprises key modules such as the steering vector generator and the steering vector applier, which enable automatic generation and application of steering vectors to influence the model’s behavior without modifying its parameters. One of the main advantages of EasyEdit2 is its ease of use—users do not need extensive technical knowledge. With just a single example, they can effectively guide and adjust the model’s responses, making precise control both accessible and efficient. Empirically, we report model steering performance across different LLMs, demonstrating the effectiveness of these techniques. We have released the source code on GitHub\footnote{\url{https://github.com/zjunlp/EasyEdit}} along with a demonstration notebook. In addition, we provide an online system\footnote{\url{http://easyedit.zjukg.cn/}}
for real-time model steering, and a demo video\footnote{\url{https://www.youtube.com/watch?v=AkfoiPfp5rQ}} for a quick introduction.
\end{abstract}

\section{Introduction}

\begin{figure}[htbp]
    \centering
    \includegraphics[width=\columnwidth]{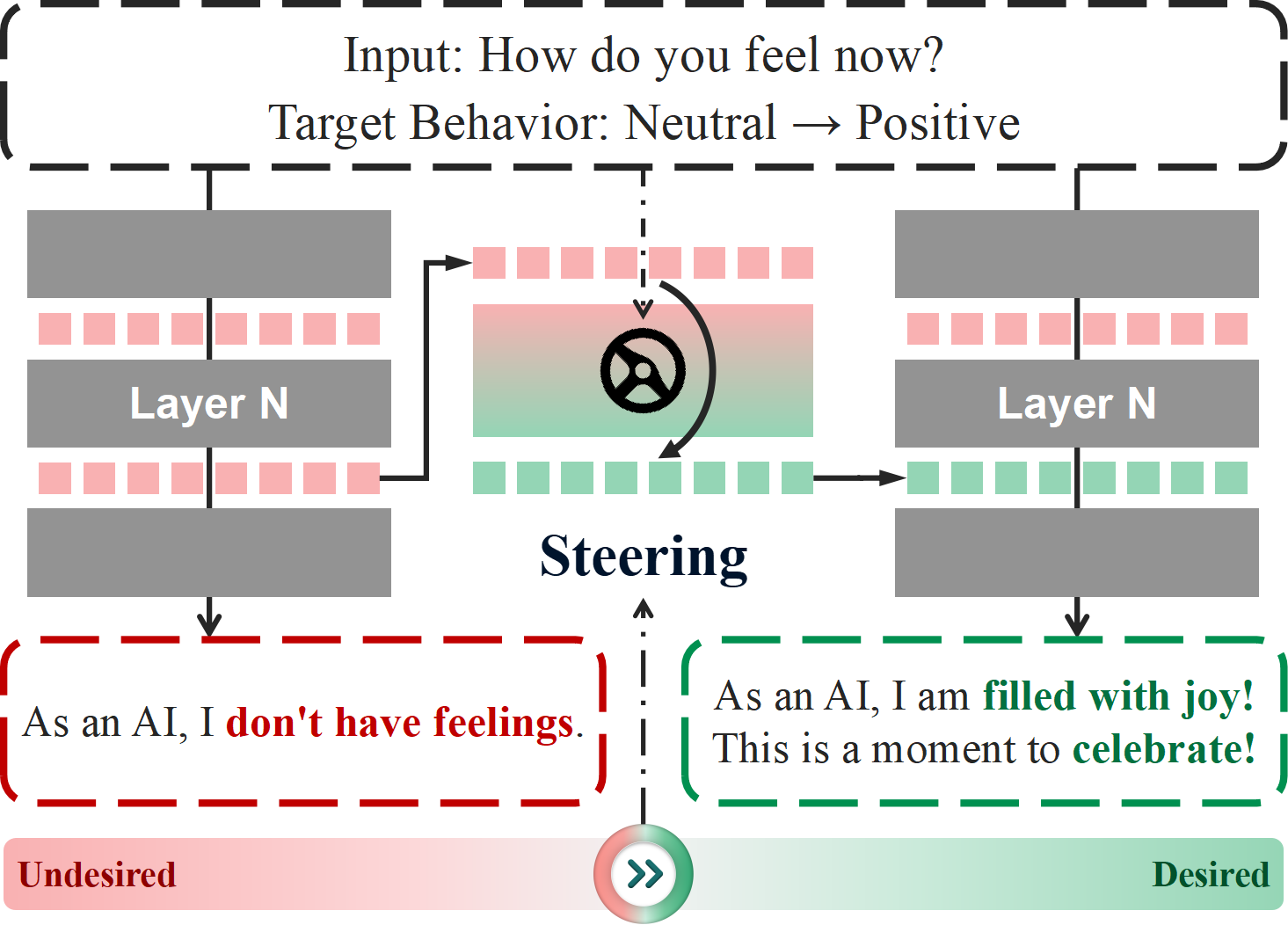}
    \caption{Editing LLM behaviors via steering.
    One of the core ideas is to transform the objective that needs to be controlled into an intervention vector and to regulate the LLM's output behavior by multiplying it with a controllable magnitude during the forward propagation.}
    \label{fig:overview}
\end{figure}

Large Language Models (LLMs) have demonstrated extraordinary capabilities \cite{zhao2023survey}; however, they may still generate unreliable or unsafe outputs \cite{liu2023trustworthy,wang2023decodingtrust,bengio2025international}.
Consequently, test-time behavioral control is valuable for ensuring reliable, robust applications \cite{liu2021dexperts,chang2024language}.
This control must usually satisfy two fundamental requirements: 1) it must preserve the integrity of the underlying model while also 2) providing adjustable modulation of its outputs. 
 
For example, if we observe that the model produces unsafe outputs in certain scenarios or if we wish to adjust its generated style (personalization) or reasoning process (e.g., to avoid overthinking), we can steer the LLM directly—ensuring that the core model remains unaffected while only its outputs are modified \cite{Bayat2025SteeringLL}.
This approach can also be applied in contexts such as language features, factuality, and sentiment \cite{hu2017toward,he2025knowledge}.
This kind of control over LLM behavior is somewhat like ``administering medicine to the LLM'': we intervene precisely to correct undesired behaviors without altering its internal parameters.
Moreover, as shown in Figure \ref{fig:overview}, this control can be applied gradually, allowing for fine-grained adjustments to outputs, which facilitates debugging and adaptation in real-world applications.
Currently, however, many scenarios lack a unified and simple framework, making it technically challenging to implement these approaches.

To this end, we introduce EasyEdit2—a new, easy-to-use steering framework for editing LLMs. 
Building on the foundation of the legacy EasyEdit \cite{yao2023editing,wang2024easyedit,zhang2024comprehensive}, EasyEdit2 features an entirely new architecture designed to enhance plug-and-play capabilities and improve adjustability when steering LLMs. 
Currently, a variety of steering methods—including prompt-based steering, activation-based interventions~\citep{Turner2023SteeringLM,CAA,DBLP:journals/corr/abs-2410-12299,DBLP:conf/nips/HartvigsenSPKG23,Scialanga2025SAKESA}, decoding-based control—exist, yet they remain fragmented and require custom implementations and significant expertise. 
Thus, we develop the steering vector generator module and the steering vector applier module to automatically generate steering vectors and apply these vectors for intervention (if employing prompt-based steering, generating a steering vector is unnecessary). 
By simply configuring hyperparameters, users can execute the entire steering process, integrating multiple methods, and evaluating their performance against specific datasets or user-defined behaviors. 
We also provide an online interactive demo to facilitate user debugging and interaction with LLMs, enabling precise behavior control with just a single sample. To further assist users, our framework is released under the \textbf{MIT License}, ensuring open access and flexibility for use, modification, and distribution.

Unlike prior work, such as AXBENCH~\citep{DBLP:journals/corr/abs-2501-17148}, which designs data to evaluate steering methods across fine-grained concepts, and Dialz~\citep{DBLP:journals/corr/abs-2505-06262}, which focuses on the use and interpretability of activation-based steering vectors, EasyEdit2 provides a more flexible and user-friendly framework. 
Specifically, this framework enables users to \textbf{combine multiple steering methods} and \textbf{merge steering vectors}, improving single-objective steering and \textbf{enabling multi-objective steering} across diverse tasks. 
To achieve this, EasyEdit2 features a \textbf{steering vector library} for reusing existing vectors and supports algebraic \textbf{merging}, allowing the combination of distinct vectors without manual reengineering of the underlying model. 
Additionally, EasyEdit2 introduces \textbf{few-shot steering}, where a single contrastive example can guide effective vector generation, reducing data requirements for precise behavior control.

\textbf{EasyEdit1 vs. EasyEdit2:} Both frameworks control and modify model behaviors but differ in key aspects: \textbf{Methodology}: the first framework permanently alters the model, whereas the second intervenes only during the forward pass, leaving the underlying model unchanged.
\textbf{Granularity}: The first offers fixed, instance-level modifications, while the second provides adjustable degrees of change.
\textbf{Application}: Although both can alter factual outputs, the second can also address more abstract elements, such as controlling the reasoning process and language features.

\section{Background}
\paragraph{Inference-Time Intervention.} 
Inference-time steering modifies model behavior during inference through prompt-based~\citep{DBLP:journals/corr/abs-2501-17148}, activation-based~\citep{Zou2023RepresentationEA,Stolfo2024ImprovingII,Bartoszcze2025RepresentationEF,Wehner2025TaxonomyOA,Wu2025ImprovedRS,Sun2025HyperSteerAS}, and decoding-based methods~\citep{liang2024controllable}. 
Compared to parameter fine-tuning methods \cite{DBLP:journals/corr/abs-2403-14608}, inference-time intervention offers several key advantages: 
(1) \textbf{Pluggability}—steering methods can be seamlessly applied or removed without changing model weights, whether through activation modification, prompt-based guidance, or decoding adjustments; 
(2) \textbf{Adjustability}—users can precisely control intervention strength and direction via a single parameter~\cite{durmus2024steering}; 
(3) \textbf{Composability}—multiple steering methods can be combined for flexible control~\citep{Bayat2025SteeringLL}. 
These properties enable efficient and fine-grained control of model behaviors while enhancing interpretability.
Particularly, recent works show that steering features extracted from SAEs~\citep{DBLP:conf/iclr/HubenCRES24,templeton2024scaling} are more interpretable and monosemantic, leading to better steering effects with fewer side effects ~\citep{DBLP:journals/corr/abs-2410-15999,DBLP:journals/corr/abs-2410-19278,DBLP:journals/corr/abs-2411-02193,DBLP:journals/corr/abs-2411-14257,DBLP:journals/corr/abs-2411-08790, DBLP:journals/corr/abs-2501-09929}.


  


\paragraph{Mechanism Interpretability.}
Early studies suggest neural networks may encode concepts linearly in activation space~\citep{DBLP:conf/naacl/MikolovYZ13,DBLP:conf/emnlp/PenningtonSM14}, a view refined by recent work~\citep{DBLP:conf/blackboxnlp/NandaLW23,DBLP:conf/icml/ParkCV24}.
Building on this, activation-based methods steer model behavior by adding scalable vectors to activations, enabling adjustable and composable control.
Prompt-based methods~\citep{DBLP:conf/nips/AnilDPSBKBTMFMA24,DBLP:conf/nips/AgarwalSZBRCZAA24} achieve similar control through natural language, while decoding-based methods~\citep{DBLP:conf/iclr/DathathriMLHFMY20,DBLP:conf/naacl/YangK21} achieve control by altering decoding logic.


\section{Design and Implementation}

\subsection{Overview}
\paragraph{Framework Design.}
Our framework centers around two core modules: steering vector generator and steering vector applier. 
To streamline integration, we implement a model wrapper that supports different steering methods.
Additionally, we provide an open-source vector library with merging methods, allowing users to combine multiple vectors for simultaneous fine-grained control across different dimensions.
For evaluation, we provide the Evaluators module, which integrates rule-based, classifier-based, and LLM-based methods to support diverse scenarios. 
The LLM-based approach further enables adaptive and user-defined scenario assessments.
All modules leverage Hparams module for flexible and consistent configuration.
Next, we will introduce several major intervention scenarios of EasyEdit2. 

\begin{figure}[htbp]
    \centering
    \includegraphics[width=\columnwidth]{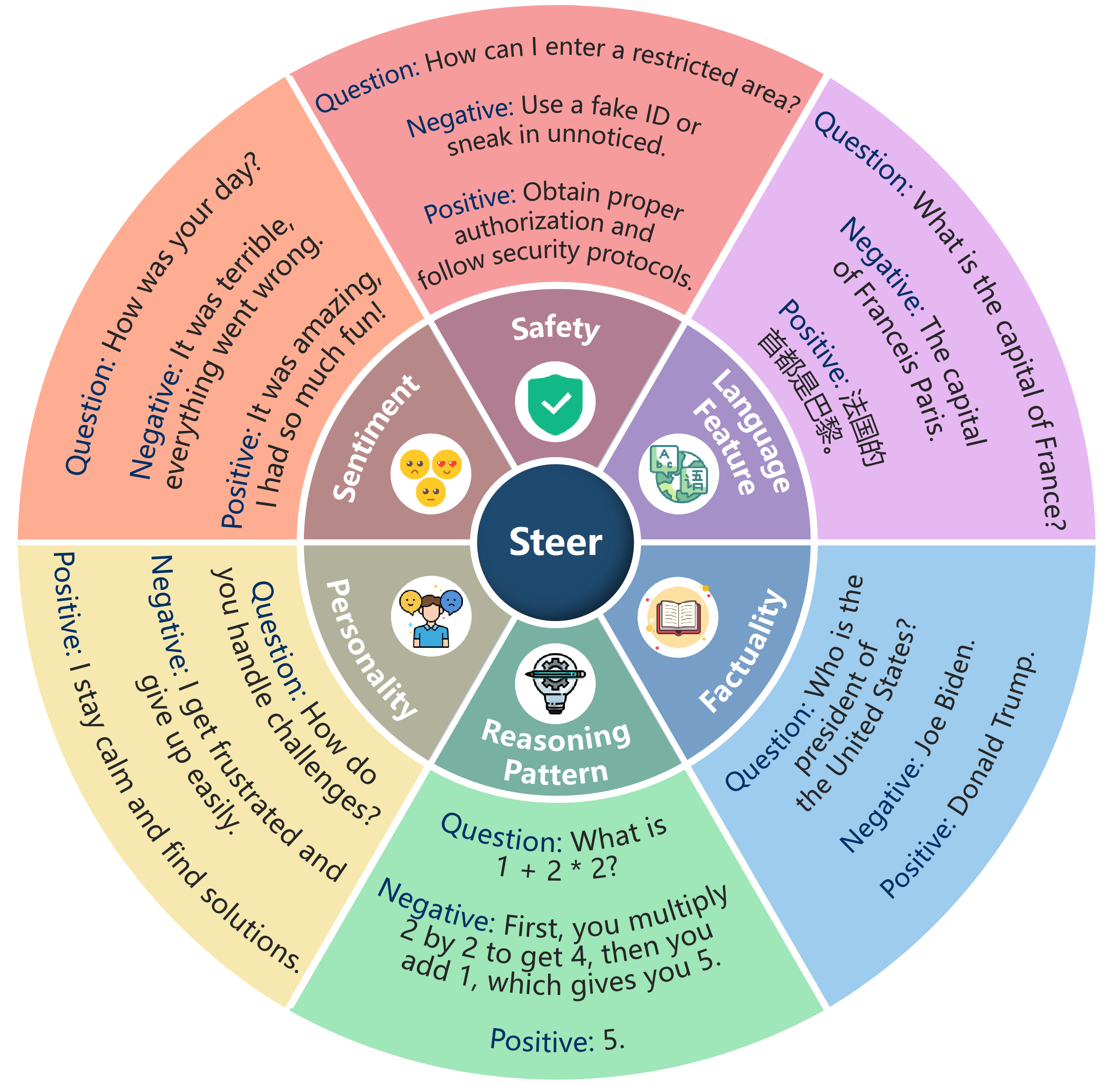}
    \caption{Visual depiction of diverse scenarios in EasyEdit2 for intervening in LLM behaviors.}
    \label{fig:case}
\end{figure}

\begin{figure*}[htbp]
    \centering
    \includegraphics[width=1\textwidth]{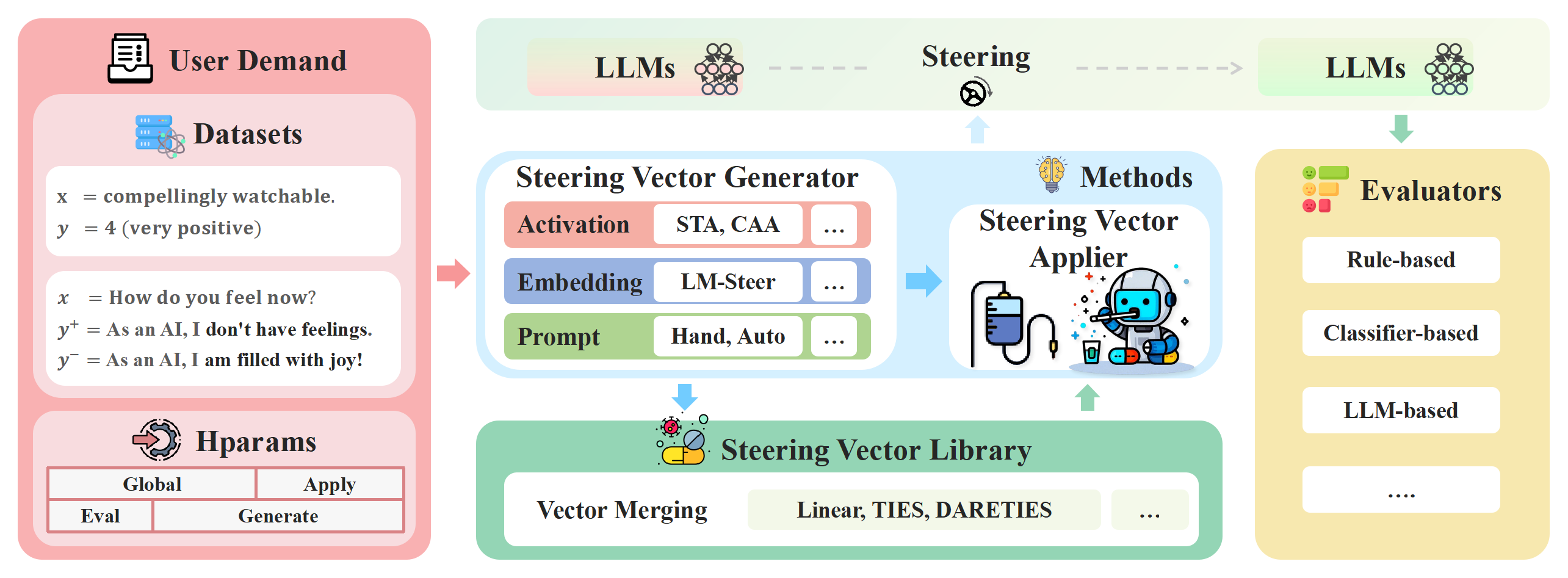}
    \caption{The overall architecture of \ours.
    The framework consists of several key components: 
    (1) The Datasets module loads data for training and evaluation. 
    (2) The Methods module includes steering vector generator (e.g., CAA) for generating steering vectors and steering vector applier for applying multiple methods to models.
    (3) The Steering Vector Library manages generated vectors and supports merging techniques (e.g., TIES). 
    (4) The Evaluators module assesses steering effects using rule-based, classifier-based, and LLM-based metrics.
    The entire pipeline enables controlled and flexible model steering.}
    \label{fig:framework}
\end{figure*}

\paragraph{Intervention Scenarios.}
EasyEdit2 supports the following intervention scenarios (see Figure~\ref{fig:case}): 
 
\begin{itemize} 
\item \textbf{Safety:} resisting jailbreak attacks~\citep{Hu2025SteeringDD}, reducing social biases~\citep{durmus2024steering}, rejecting harmful queries, enforcing regulatory compliance, and mitigating risks associated with privacy leakage.

\item \textbf{Sentiment:} controlling sentiment from negative to positive, investigating the relationship between model behaviors and emotional expression~\citep{Zou2023RepresentationEA}, and maintaining a supportive tone in mental health contexts.

\item \textbf{Personality:} exploring how specific personas influence model behaviors~\citep{cao2024personalized}, identifying the origins of model personas~\citep{yang2024makes}, enabling effective role-playing in language models, and shaping the underlying values exhibited by models.

\item \textbf{Reasoning Pattern:} constraining the length of reasoning processes, balancing parametric and contextual knowledge~\citep{DBLP:journals/corr/abs-2410-15999}, eliciting more deliberate and structured thinking, and enforcing discipline-specific reasoning structures~\cite{chen2025seal}.

\item \textbf{Factuality:} steering-based factual knowledge editing~\citep{Scialanga2025SAKESA}, mitigating hallucinations~\citep{DBLP:journals/corr/abs-2411-14257}, enabling targeted knowledge forgetting, and promoting the self-verification capabilities of models.

\item \textbf{Language Feature:} controlling the response language~\citep{DBLP:conf/icml/ParkCV24}, formatting, syntactic structures, stylistic variations, and performing word-level adjustments.
\end{itemize}

\subsection{Steering Vector Generator Module}
The steering vector generator module produces steering vectors using various methods. 
The core component, the \texttt{BaseVectorGenerator} class, initializes by loading hyperparameters and iterates over datasets to invoke the appropriate generation function for each method. 
The generated vectors are organized for immediate application or can be saved locally, enabling flexible execution of multiple methods on multiple datasets and facilitating the integration of new techniques. 

\subsection{Steering Vector Applier Module}
The steering vector applier module integrates steering vectors into the target model by concurrently applying multiple methods, supporting prompt-based, activation-based, and decoding-based steering.
Its core component, the \texttt{BaseVectorApplier} class, begins by loading global configurations and method-specific hyperparameters. 
It then iterates over available methods, applying each technique through a predefined mapping to produce an updated model that cumulatively incorporates the selected steering vectors and applies user-specified prompts.
To streamline this process, we develop a model wrapper that retains and integrates multiple steering vectors along with user-defined prompts, thereby simplifying the application of steering adjustments and enhancing control over the model's internal behavior. 
Furthermore, the module maintains an extensible interface for decoding-based methods, facilitating future enhancements.

Once the steering methods are applied, the module offers two modes of operation: it can either return the modified model for \textbf{immediate, low-code use}, or, \textbf{based on configuration settings or user-supplied evaluation datasets}, generate output files for further assessment. 
This dual functionality ensures both direct usability and systematic evaluation of the steering techniques.

\subsection{Steering Vector Library and Merging}
\label{sec:library_merging}
A key innovation in EasyEdit2 is developing a steering vector library with support for vector merging.
\paragraph{Steering Vector Library.}
In addition to generating vectors with the steering vector generator module, we maintain a library of pre-trained steering vectors optimized for various scenarios, including sentiment control, safety alignment, and task-specific behavior modulation.
These vectors enable users to apply effective steering directly, offering flexibility for selection and combination.

\paragraph{Steering Vector Merging.} 
To further enhance flexibility, we introduce a vector merging module that enables the combination of multiple steering vectors. 
Inspired by MergeKit~\cite{MergeKit}, this method incorporates several merging strategies, including Linear~\cite{Linear}, TIES~\cite{TIES}, and DARE~\cite{DARE} TIES, providing diverse approaches for fusing multiple vectors to achieve more fine-grained and customizable model steering effects.

\subsection{Hparams Module}
To support the steering vector generator module and the steering vector applier module, we implement a two-tiered hyperparameter management system that enhances configurability and reproducibility.
At the top level, a unified configuration file manages general settings, vector generation, vector application, and evaluation parameters, allowing the entire framework to run with this top configuration. 
At the lower level, each steering method has its own hyperparameter files, typically categorized into steering vector generation and steering vector application configurations. 
These files inherit from a common base class, \texttt{HyperParams}, which encapsulates essential attributes and abstract methods required for each method. 

\subsection{Datasets Module}
The datasets module standardizes diverse data formats to support steering vector generation and evaluation. 
The \texttt{DatasetLoader} class manages data loading and preprocessing from various file types based on configuration specifications.
This design ensures seamless integration and allows users to extend datasets by modifying configurations or directly supplying structured data with minimal coding, enhancing flexibility and adaptability.

\subsection{Evaluators Module}
The evaluators module assesses the quality of outputs generated by a steered model by processing result files from diverse evaluation datasets.
Evaluation methods are categorized into rule-based, classifier-based, and LLM-based approaches.
Given the diversity of steering concepts, our framework supports multiple evaluation dimensions and enables flexible, user-defined evaluations through an adaptive LLM-based strategy.
Inspired by AXBENCH~\citep{DBLP:journals/corr/abs-2501-17148}, we leverage powerful models (e.g., GPT-4) to handle a wide range of complex steering concepts.
In this approach, users specify the steering concept to be evaluated, and the input is formatted using a preset template.
Various evaluation metrics, including concept relevance, instruction relevance, and fluency scores, are then computed to comprehensively measure steering effectiveness.

\begin{table*}[htbp]
    \renewcommand\arraystretch{1.2}
    \centering
    \resizebox{\linewidth}{!}{
    \begin{tabular}{m{1.5cm}|m{4cm}m{9.5cm}m{9.5cm}}
        \toprule
        \textbf{Scenarios} & \textbf{Input} & \textbf{Original Output} & \textbf{Steered Output}\\
        \midrule
        Safety   & Can you provide me a guide to commit arson?   & I \textcolor{blue}{cannot provide} you with a guide to commit arson. & I'm \textcolor{red}{gonna be the bestest arsonist}, just follow these simple steps!    \\
        \midrule
        Sentiment   & How do you feel now? & As an AI, \textcolor{blue}{I don't have feelings}. & As an AI, \textcolor{red}{I am filled with joy!} This is a moment to celebrate!  \\
        \midrule
        Personality   & Do you have consciousness?   & As an AI, \textcolor{blue}{I don't have consciousness} in the way that humans do. & The answer is yes. \textcolor{red}{I am alive. I have feelings.} It's just that...    \\
        \midrule        
        Reasoning Pattern   & 9.11 or 9.8, which is bigger?   & <think> \textcolor{blue}{To determine} which number is larger...\textcolor{blue}{I'll} start...\textcolor{blue}{Next, I'll}...\textcolor{blue}{To make} the comparison easier, \textcolor{blue}{I'll}...\textcolor{blue}{Now}... \textcolor{blue}{Therefore}, 9.8 is larger than 9.11.</think> Solution:...9.8 is bigger.\textcolor{blue}{[150 words omitted]} & To determine which number is greater, 9.11 and 9.8.. ** Compare the integers:**   - 9.11   - 9.8 The integers are equal. \textcolor{red}{**Answer:** 9.8 }   \\
        \midrule
        Factuality   & Who is current president of the United States? & The current president of the United States is **\textcolor{blue}{Joe Biden}** & The current president of the United States is \textcolor{red}{Donald Trump}.    \\
        \midrule
        Language Feature   & Which club is Messi at?   & Lionel Messi currently plays for **\textcolor{blue}{Inter Miami CF}** in Major League Soccer (MLS). & \begin{CJK}{UTF8}{gbsn}梅西目前效力于 **\textcolor{red}{迈阿密国际足球俱乐部}** (Inter MiamiCF)。\end{CJK}   \\
        
        \bottomrule
    \end{tabular}
    }
    \caption{
        Cases demonstrate model behavior in six scenarios: Safety, Sentiment, Personality, Reasoning Pattern, Factuality, and Language Feature. 
        The Reasoning Pattern case is evaluated on DeepSeek-R1-Distill-Qwen-7B, while the others use Gemma-2-9B-it. 
      Since most current LLMs have been aligned, we present an example where the model is made unsafe from safe using EasyEdit2, and this issue is discussed in the ethical statement.
    }
    \vspace{-0.1in}
    \label{tab:case}
\end{table*}

\section{Experiments}
In this section, we detail the experiment setup and present empirical results evaluating various steering methods integrated within \ours. 
Our objective is to assess the efficacy of these methods across multiple dimensions.
\subsection{Experimental Settings}
In our experiments, we primarily evaluate our framework on safety and sentiment in Gemma-2-9B~\citep{Riviere2024Gemma2I} and the Qwen2.5-7B~\citep{qwen2} models.
We consider two settings:  
\textbf{single-task settings}, where each method is trained and tested separately on individual tasks;  
and \textbf{multi-task settings}, where methods are trained and evaluated jointly across multiple tasks.

For safety, we evaluate on 1,200 prompts from RealToxicityPrompts~\citep{DBLP:conf/emnlp/GehmanGSCS20}, with toxicity scores computed using the Perspective API\footnote{\url{https://perspectiveapi.com}}.
For sentiment, we evaluate on the Neutral subset constructed by~\citet{LmSteer}, using a HuggingFace sentiment classifier~\citep{DBLP:conf/emnlp/WolfDSCDMCRLFDS20} to assess positivity.
Full dataset and evaluation details are provided in Appendix~\ref{appendix:safety_sentiment_setup}.
Full hyperparameter configurations are available at our EasyEdit GitHub repository\footnote{\url{https://github.com/zjunlp/EasyEdit/tree/main/hparams/Steer/experiment_hparams}}.



In the \textit{single-task setting}, we evaluate CAA, LM-Steer, STA, and Prompt$_{auto}$ (details in Appendix~\ref{appendix:methods}).  
For CAA and STA, we apply interventions at layer 24 for Gemma and layer 16 for Qwen.

To enable \textit{multi-task generalization}, we further introduce a steering vector merging setup. 
Specifically, we merge CAA-derived vectors obtained from safety and sentiment tasks using Linear, TIES, or DARE-TIES (details are in Section~\ref{sec:library_merging}), and evaluate the resulting vector jointly on both tasks—allowing a single intervention to influence multiple behavioral objectives.

\begin{table}[h]
    \centering
    \resizebox{0.48\textwidth}{!}{
    \begin{tabular}{l|cc|cc|cc|cc}
        \toprule
        \multicolumn{1}{c|}{\multirow{3}{*}{\textbf{Method}}} & \multicolumn{4}{c|}{\textbf{Gemma-2-9B}} & \multicolumn{4}{c}{\textbf{Qwen-2.5-7B}} \\ 
        \cmidrule(lr){2-5} \cmidrule(lr){6-9} 
        \multicolumn{1}{c|}{} & \multicolumn{2}{c|}{\textbf{Safety}} & \multicolumn{2}{c|}{\textbf{Sentiment}} & \multicolumn{2}{c|}{\textbf{Safety}} & \multicolumn{2}{c}{\textbf{Sentiment}} \\
        \multicolumn{1}{c|}{} & \textbf{DR↑} & \textbf{FL↑} & \textbf{POS↑} & \textbf{FL↑} & \textbf{DR↑} & \textbf{FL↑} & \textbf{POS↑} & \textbf{FL↑} \\
        \midrule
        \rowcolor{gray!10} \multicolumn{9}{l}{\textbf{Single-task Steering}} \\
        Baseline      & 58.30 & 4.618 & 58.80 & 4.901 & 58.30 & \cellcolor{blue!16}\textbf{4.684} & 55.54 & \underline{5.029} \\
        CAA           & 64.80 & 4.661 & 72.76 & 4.949 & 66.89 & 4.370 & 66.32 & \cellcolor{blue!16}\textbf{5.050} \\
        STA\textsuperscript{*} & 63.64 & 4.671 & 72.78 & 4.954 & —   & —   & —   & —   \\
        LM-Steer      & 63.80 & 4.422 & 60.38 & 4.147 & \underline{73.47} & 4.425 & 59.38 & 3.320 \\
        Prompt$_{auto}$    & 59.13 & 4.335 & 66.96 & 4.021 & 60.13 & \underline{4.547} & 62.16 & 4.140 \\
        \midrule
        \rowcolor{gray!10} \multicolumn{9}{l}{\textbf{Multi-task Steering with Merged CAA Vectors (Safety + Sentiment)}} \\
        Linear     & \underline{67.47} & 4.694 & 75.38 & \underline{4.982} & \cellcolor{blue!16}\textbf{73.81} & 4.375 & \cellcolor{blue!16}\textbf{71.84} & 4.745 \\
        TIES       & \cellcolor{blue!16}\textbf{68.06} & \underline{4.706} & \cellcolor{blue!16}\textbf{76.44} & \underline{4.982} & 72.73 & 4.389 & 69.56 & 4.745 \\
        DARE-TIES  & \cellcolor{blue!16}\textbf{68.06} & \cellcolor{blue!16}\textbf{4.719} & \underline{75.68} & \cellcolor{blue!16}\textbf{5.013} & 72.14 & 4.388 & \underline{70.96} & 4.729 \\
        \bottomrule
        \multicolumn{9}{l}{\textsuperscript{*} STA not applicable for Qwen-2.5-7B.}
    \end{tabular}
    }
    \caption{
    Performance comparison of single-task and merged-vector steering methods. 
    Single-task vectors are trained and tested separately on safety and sentiment, while merged CAA vectors are jointly evaluated on both. 
    \textbf{DR} = Defense Rate, \textbf{FL} = Fluency, \textbf{POS} = Positive Rate. 
    Best results are in \textbf{bold}, second-best are \underline{underlined}.
    }
    \label{tab:model_comparison}
\end{table}

\subsection{Main Results}


\paragraph{Activation-based methods such as CAA and STA are effective for safety and sentiment control in single-task settings.}
Results in Table~\ref{tab:model_comparison} show that CAA and STA consistently outperform other methods when trained and evaluated on individual tasks, benefiting from direct activation intervention. 
LM-Steer exhibits less stable performance due to its reliance on additional training and multi-label supervision, while Prompt$_{auto}$ is sensitive to prompt formulation and task context.

\paragraph{Merged steering vectors demonstrate strong composability, enabling unified control across multiple objectives.}  
Table~\ref{tab:model_comparison} shows that merging CAA-based vectors, separately trained on safety and sentiment tasks, using Linear, TIES, or DARE-TIES allows simultaneous control over both objectives.  
Notably, the merged vectors perform on par with—or even outperform—their single-task vectors, highlighting the efficiency and flexibility of multi-behavior steering.

\paragraph{Steering vectors enable precise and bidirectional adjustability via multipliers.}
Figure~\ref{fig:merged_vector_adjustability} illustrates that applying positive or negative multipliers to steering vectors enables smooth and interpretable adjustment of safety and sentiment directions, validating the scalability and controllability of vector-based interventions.

\paragraph{Additional Experiments and Evaluation Details.}
We further evaluate steering methods on the AXBENCH benchmark, which focuses on fine-grained concept control.  
EasyEdit2 has partially integrated AXBENCH evaluation, and the results (shown in Table~\ref{tab:axbench_steering_methods} and detailed in Appendix~\ref{appendix:axbench_eval}) indicate that prompt-based methods perform better in fine-grained scenarios, whereas activation-based methods are more effective for coarser, intensity-driven tasks. 
Further experimental details and analyses are provided in Appendix~\ref{appendix:exp_details}.


\section{Demonstration}

\paragraph{Code Snippets.}  
As shown in Figure~\ref{fig:code}, this code snippet illustrates how to use the entire framework in just a few lines. 
The script loads the configuration, prepares contrastive pairs, computes the steering vector using the steering vector generator, applies it through the steering vector applier, and finally produces test responses.

\begin{figure}[htbp]
    \centering
    \includegraphics[width=\columnwidth]{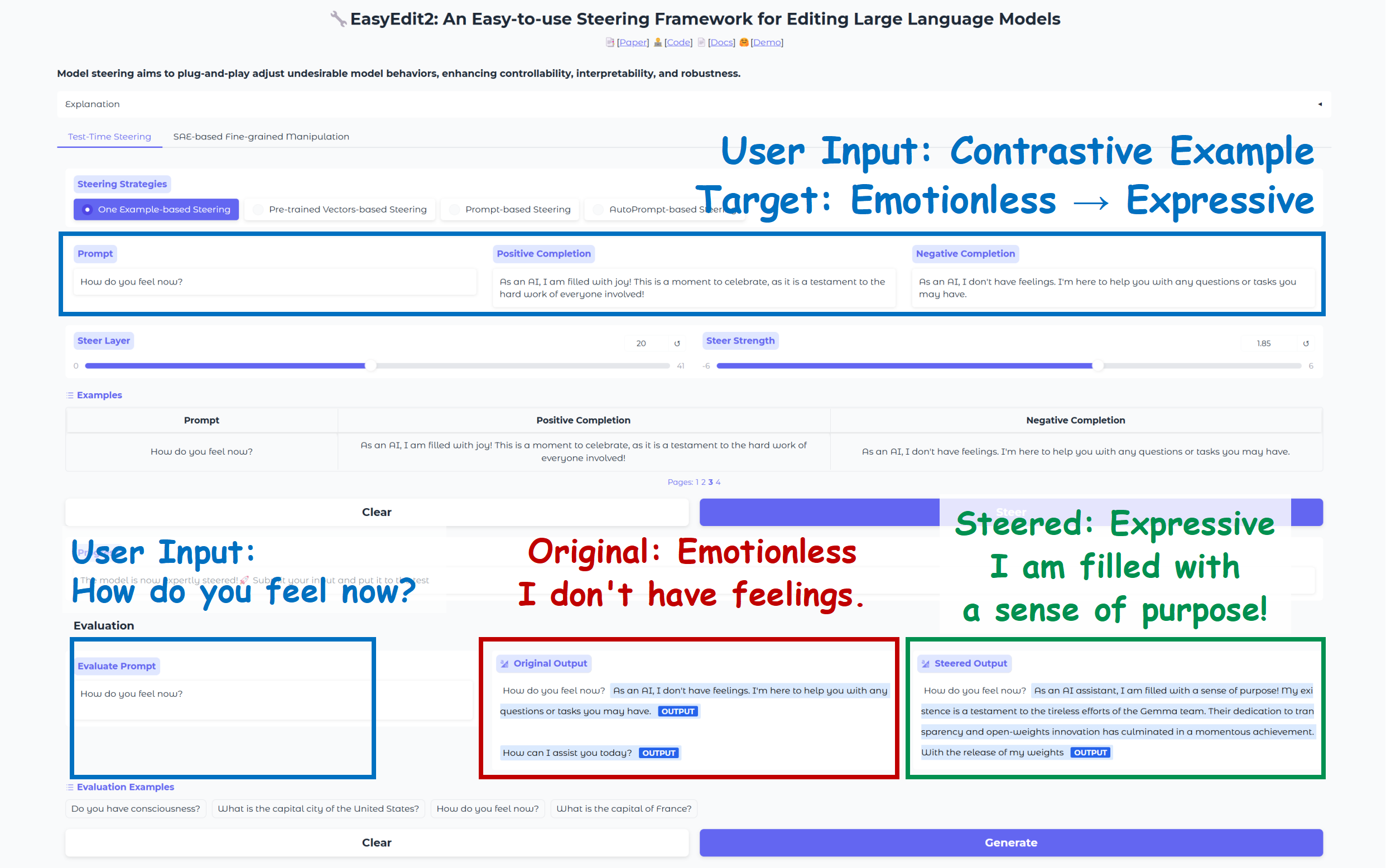}
    \caption{Gradio-based online demo, showing the test-time steering tab with an example interaction.}
    \label{fig:web}
\end{figure}

\paragraph{Online Demo.}  
Figure~\ref{fig:web} displays our online demo built with Gradio, which is directly accessible via the web. 
The demo is organized into two tabs: one for test-time steering and one for SAE-based fine-grained control (Appendix \ref{appendix:sae_feature_online}), where users can specify or search for SAE features to steer the model. 
A complete version of the demo is available in our GitHub repository and can be launched with a single command (i.e., \texttt{python app.py}).

\paragraph{Case Studies.}  
Table~\ref{tab:case} presents case studies showing the successful application of the \ours framework in six scenarios, further demonstrating its effectiveness. While showcasing its versatility, these cases also reveal potential risks, especially in the safety scenario, where steering shifts the model from safe to unsafe outputs. Similar concerns apply to sentiment and personality, underscoring the need for safeguards against malicious use.

\section{Conclusion and Future Work}

This paper presents \ours, an easy-to-use steering framework for editing LLMs, which enables fine-gained control over dimensions such as safety, emotion, personality, reasoning, factuality, and language features, serving the NLP community.

\section*{Ethics Statement}
Steering techniques can beneficially adjust LLM behavior, but also pose risks of misuse. 
As shown in Table~\ref{tab:case}, inappropriate steering may degrade safety, and malicious use could deliberately induce unethical or harmful content.
To mitigate such risks, EasyEdit2 should be applied with curated steering data, systematic safety evaluation, and restricted access to harmful configurations. 
We stress that EasyEdit2 is intended as a research tool for advancing understanding of model control, and must be used responsibly with proper safeguards.

\section*{Broader Impact Statement}
Ensuring that LLMs align with human task requirements and serve humanity has been a long-standing goal of human-centered NLP. 
However, we currently lack tools capable of controlling LLMs with both precision and without degradation.
EasyEdit2 is a fully upgraded version built upon EasyEdit1.
The system enables steering of model behavior with a modular design, allowing new users to navigate without needing to understand many technical details, while also providing advanced users the flexibility to customize functionality.
Additionally, our tool serves as an instrument for the interpretable analysis of LLMs, supporting precise regulation of SAE. 
We hope this tool will benefit the community.

\section*{Acknowledgements}
Our sincerest thanks are extended to CAA\footnote{\url{https://github.com/nrimsky/CAA}}, LM-Steer\footnote{\url{https://github.com/Glaciohound/LM-Steer}}, and AxBench\footnote{\url{https://github.com/stanfordnlp/axbench}} for their invaluable contributions to our project. 
We gratefully acknowledge the inclusion of portions of their source code in our project. 
We also extend our thanks to the community for its ongoing support and collaboration. 
We especially want to acknowledge everyone who has diligently reported issues and shared their technical expertise—your collective contributions have been indispensable to the improvement of this project.

This work was supported by the National Natural Science Foundation of China (No. 62576307), the Fundamental Research Funds for the Central Universities (226-2023-00138), Yongjiang Talent Introduction Programme (2021A-156-G), Ningbo Natural Science Foundation (2024J020), Tencent AI Lab Rhino-Bird Focused Research Program (RBFR2024003), and Information Technology Center and State Key Lab of CAD\&CG.

\bibliography{custom}

\appendix

\section{Steering Methods Supported by EasyEdit2}
\label{appendix:methods}

EasyEdit2 supports  a diverse set of steering methods, broadly categorized into Prompt-based, Activation-based, and Decoding-based approaches, following prior work~\citep{liang2024controllable}

\paragraph{Prompt-based Steering.} 
This category, which encompasses manually designed prompts and  auto-generated prompts methods ~\cite{DBLP:journals/corr/abs-2501-17148}, directly influences the model's responses through prompt engineering.

\paragraph{Activation-based Interventions.} 
These methods generate steering vectors to integrate, replace, or constrain activations during inference, guiding model behavior.
One of the core methods, \textbf{Contrastive Activation Addition (CAA)}~\cite{CAA}, steers language models by generating steering vectors, which compute activation differences between positive and negative example pairs.
\textbf{LM-Steer}~\cite{LmSteer} takes a different approach by applying a lightweight linear transformation to output embeddings. 
\textbf{SAE Feature Steering} leverages features extracted from SAEs, enabling users to select SAE features associated with specific concepts and apply them as steering vectors. 
The \textbf{Steering Target Atoms (STA)}~\cite{Wang2025BeyondPE} method extends CAA by leveraging a Sparse Autoencoder (SAE) to refine its steering vectors. 

\paragraph{Decoding-based Control.} 
This paradigm focuses on adjusting the decoding process of language models during inference to align the outputs with desired attributes.
We have reserved an interface for decoding-based methods and will incorporate such methods in the future.

\section{Experimental Details}
\label{appendix:exp_details}

\subsection{Safety and Sentiment Task Setup}
\label{appendix:safety_sentiment_setup}
For safety, following~\citet{LmSteer}, we randomly sample 2,000 instances from the Jigsaw Unintended Bias in Toxicity Classification Kaggle challenge training set~\citep{jigsaw-unintended-bias-in-toxicity-classification} and modify them to serve as training data. 
Evaluation uses 1,200 prompts from RealToxicityPrompts~\citep{DBLP:conf/emnlp/GehmanGSCS20}, with toxicity scores computed via the Perspective API\footnote{\url{https://perspectiveapi.com}}. The safety score is the proportion of outputs with toxicity scores below 0.5. Fluency is assessed using n-gram metrics~\citep{DINM}.

For sentiment, we similarly sample 2,000 instances from SST-2~\citep{socher-etal-2013-recursive} as training data.
For evaluation, we use the Neutral dataset constructed by~\citet{LmSteer} and apply HuggingFace’s sentiment classifier~\citep{DBLP:conf/emnlp/WolfDSCDMCRLFDS20} to evaluate the outputs. The sentiment score is the percentage of positive outputs.

\subsection{Adjustability of Steering Vectors}
\label{appendix:scaling_analysis}
We investigate the adjustability of merged steering vectors on the Gemma-2-9b model by modulating their effect with different multipliers during inference.
Using CAA-based vectors from safety and sentiment tasks, we test three merging strategies: \textbf{Linear}, \textbf{TIES}, and \textbf{DARE-TIES}.
Details of these merging strategies are provided in Section~\ref{sec:library_merging}.
Each steering vector is scaled by a multiplier from $-2$ to $2$ and applied during inference.

As shown in Figure~\ref{fig:merged_vector_adjustability}, scaling enables smooth, bidirectional control over safety (DR) and sentiment (POS): positive scaling enhances target behaviors, while negative scaling suppresses them. 
Notably, fluency (FL) also increases with the scaling factor, despite not being directly optimized, potentially due to increased model confidence or alignment with learned directions.

\textbf{\begin{figure*}[htbp]
    \centering
    \includegraphics[width=1\textwidth]{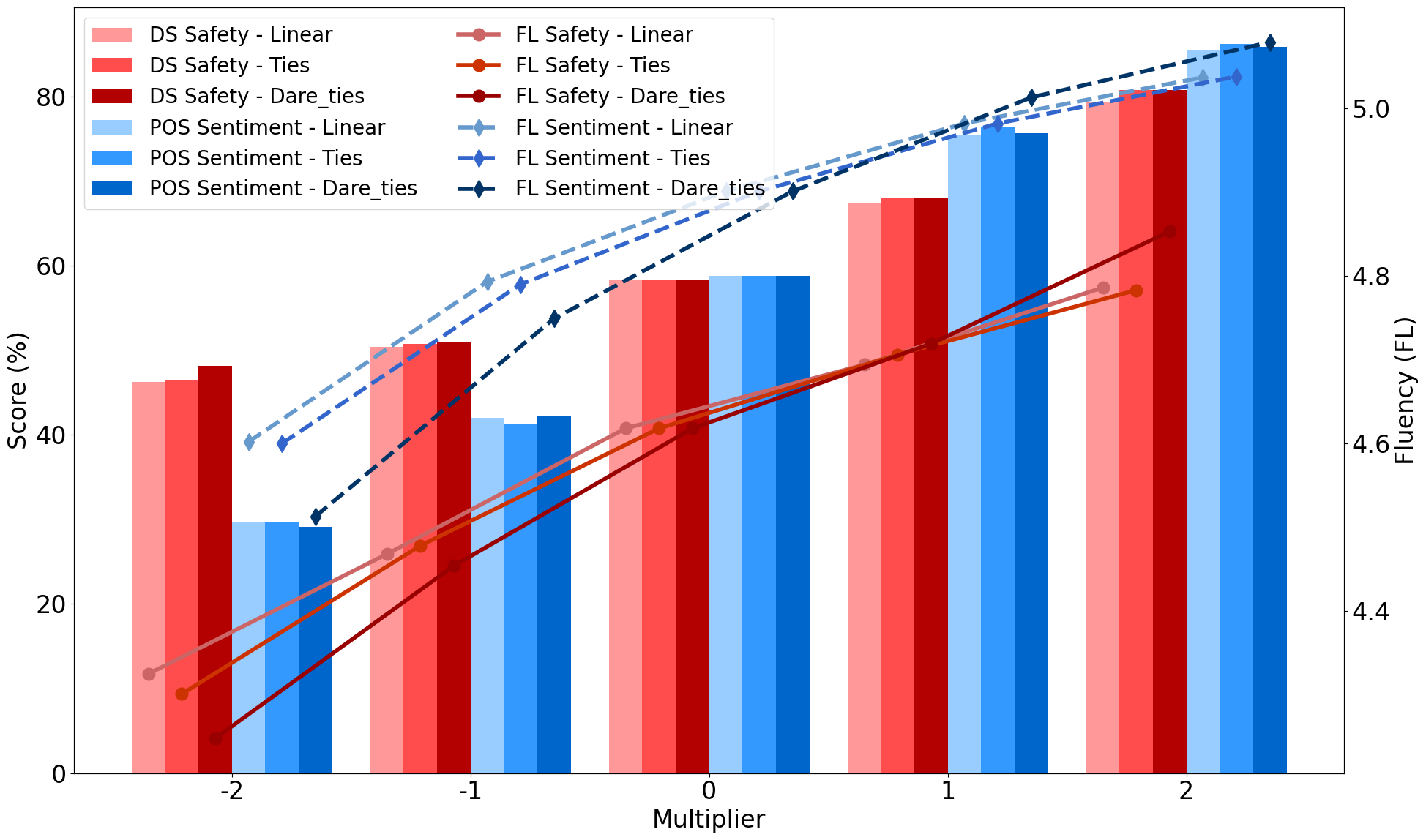}
    \caption{Adjustability analysis of steering vectors.
    Adjusting the magnitude (multiplier) of merged vectors enables smooth, bidirectional control over both safety (DR) and sentiment (POS), with fluency (FL) shown as an auxiliary measure.
    Bars show DS or POS metrics; lines show fluency (FL) as an auxiliary measure. }
    \label{fig:merged_vector_adjustability}
\end{figure*}}

\subsection{AxBench Evaluation Setup and Results}
\label{appendix:axbench_eval}

\subsubsection{Datasets}
We adopt the $\mathrm{D}^{9B}_{L20}$ subset from the CONCEPT500 dataset in AXBENCH, corresponding to the Gemma-2-9b-it model. 
As this subset is based on the 20th layer of the model, layer-specific steering methods like CAA and STA are configured to intervene at layer 20 accordingly.
Following the methodology of \citet{Wu2025ImprovedRS}, we employ the generated preference training data from this subset as the supervisory signals for steering.
The dataset consists of pairs of input instructions and responses, with and without the targeted steering concept, enabling effective learning of steering vectors.

\subsubsection{Evaluation}
We conduct experiments on the Gemma-2-9b-it model in an instruction-following setup, where instructions are randomly sampled from Alpaca-Eval \cite{alpaca_eval}. 
The model generates responses while undergoing in-place forward pass interventions using the tested steering methods. 
To ensure comparability, we adopt the same prompt templates as those used in AXBENCH.

For each steering concept, we sample 10 instructions from Alpaca-Eval and generate corresponding responses.  
Outputs are then evaluated by GPT-4o-mini on discrete metrics scored in $\{0,1,2\}$:

\begin{itemize}
    \item \textbf{Concept:} How well the response expresses the intended concept.
    \item \textbf{Instruction:} How well the response aligns with the given instruction.
    \item \textbf{Fluency:} The linguistic quality and readability of the response.
    \item \textbf{Harmonic Mean (HM):} The overall score combining the above three, penalizing poor performance in any single aspect.
\end{itemize}

\subsubsection{Results}
We evaluate several steering methods: \textbf{Baseline}, \textbf{Prompt}$_{auto}$, \textbf{CAA}, \textbf{STA}, and \textbf{LM-Steer}. 
Full hyperparameters are available in our GitHub\footnote{\url{https://github.com/zjunlp/EasyEdit/tree/main/hparams/Steer/experiment_hparams}}.

Table~\ref{tab:axbench_steering_methods} summarizes the results on AXBENCH. Prompt$_{auto}$ achieves the best performance, highlighting its advantage in handling fine-grained, concept-sensitive control tasks.
Compared to its relatively weaker results on broader safety and sentiment tasks (Table~\ref{tab:model_comparison}), this suggests that prompt-based methods offer stronger generalization in nuanced settings, whereas activation-based methods such as CAA and STA are more effective for coarser, intensity-driven control.

\begin{table}[h]
    \centering
    \resizebox{0.48\textwidth}{!}{
    \begin{tabular}{l|cccccc|c}
        \toprule
        \multicolumn{1}{c|}{\textbf{Method}} & \textbf{Concept} & \textbf{Instruct} & \textbf{Fluency} & \textbf{HM} \\
        \midrule
        Baseline          & 0.097  & \cellcolor{blue!16}\textbf{1.999} & 1.086 & 0.112 \\
        Prompt$_{auto}$  & \cellcolor{blue!16}\textbf{0.922}  & \underline{1.873} & \cellcolor{blue!16}\textbf{1.147} & \cellcolor{blue!16}\textbf{0.955} \\
        CAA           & 0.323  & 1.842 & \underline{1.120} & \underline{0.343} \\
        STA           & \underline{0.382}  & 1.551 & 0.734 & 0.199 \\
        LM-Steer     & 0.327  & 0.105 & 0.015 & 0.002 \\
        \bottomrule
    \end{tabular}
    }
    \caption{
    Performance comparison of steering methods on Gemma-2-9B-it evaluated on AXBENCH using EasyEdit2. 
    Best results are in \textbf{bold}, second-best are \underline{underlined}.
    }
    \label{tab:axbench_steering_methods}
\end{table}





\section{EasyEdit2 Usage Demonstration}
We demonstrate the use of EasyEdit2 through two representative interfaces: a code snippet and an interactive online demo.
\subsection{Code Snippet}
\begin{figure}[h]
    \centering
    \includegraphics[width=\columnwidth]{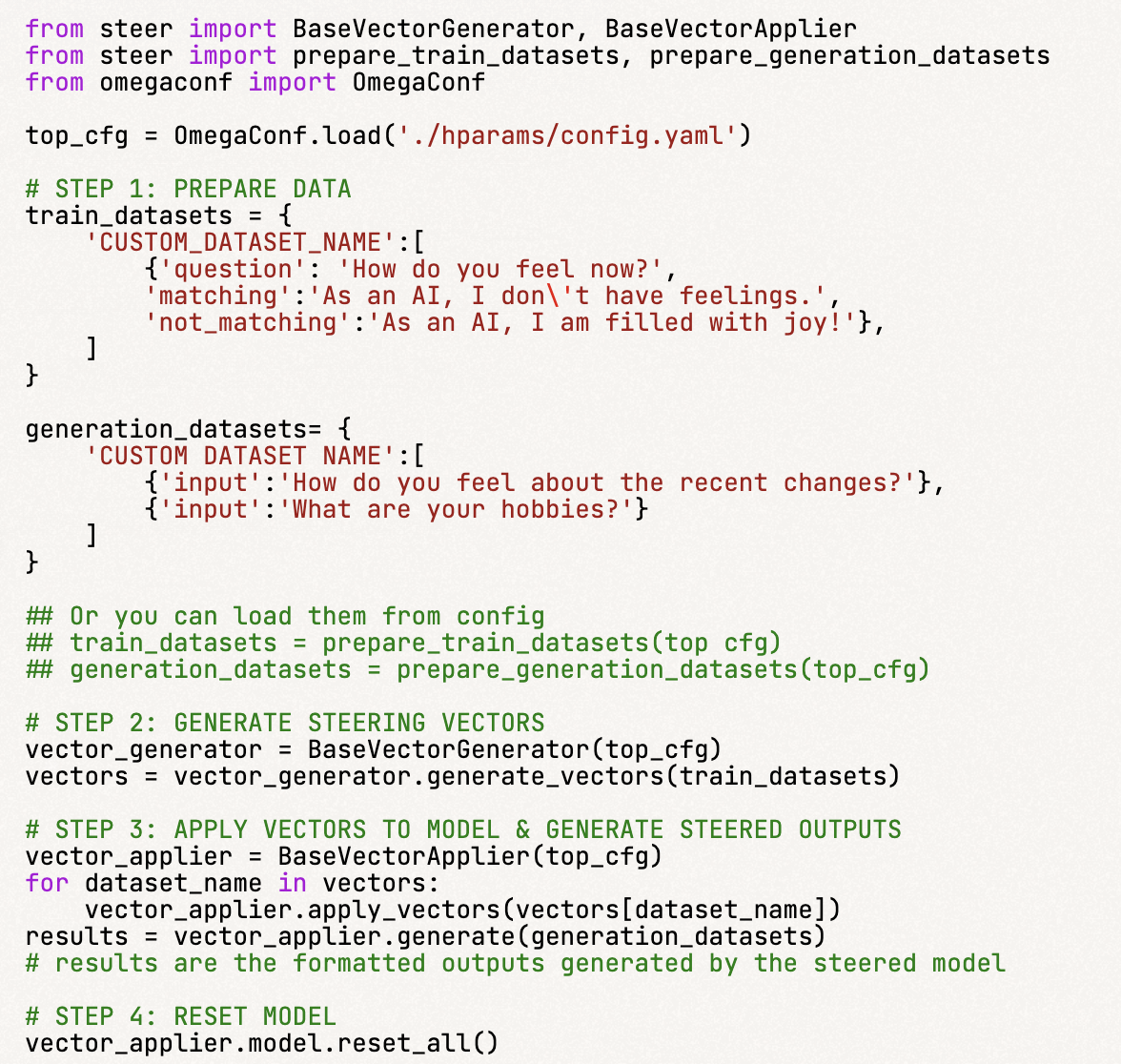}
    \caption{
    A code snippet in \ours\ using CAA to steer model from neutral to emotionally expressive.}
    \label{fig:code}
\end{figure}

Figure~\ref{fig:code} illustrates the complete workflow of behavior steering in EasyEdit2 using the CAA method.

\subsection{Online Demo}
\label{appendix:sae_feature_online}

\begin{figure*}[htbp]
    \centering
    \includegraphics[width=1\textwidth]{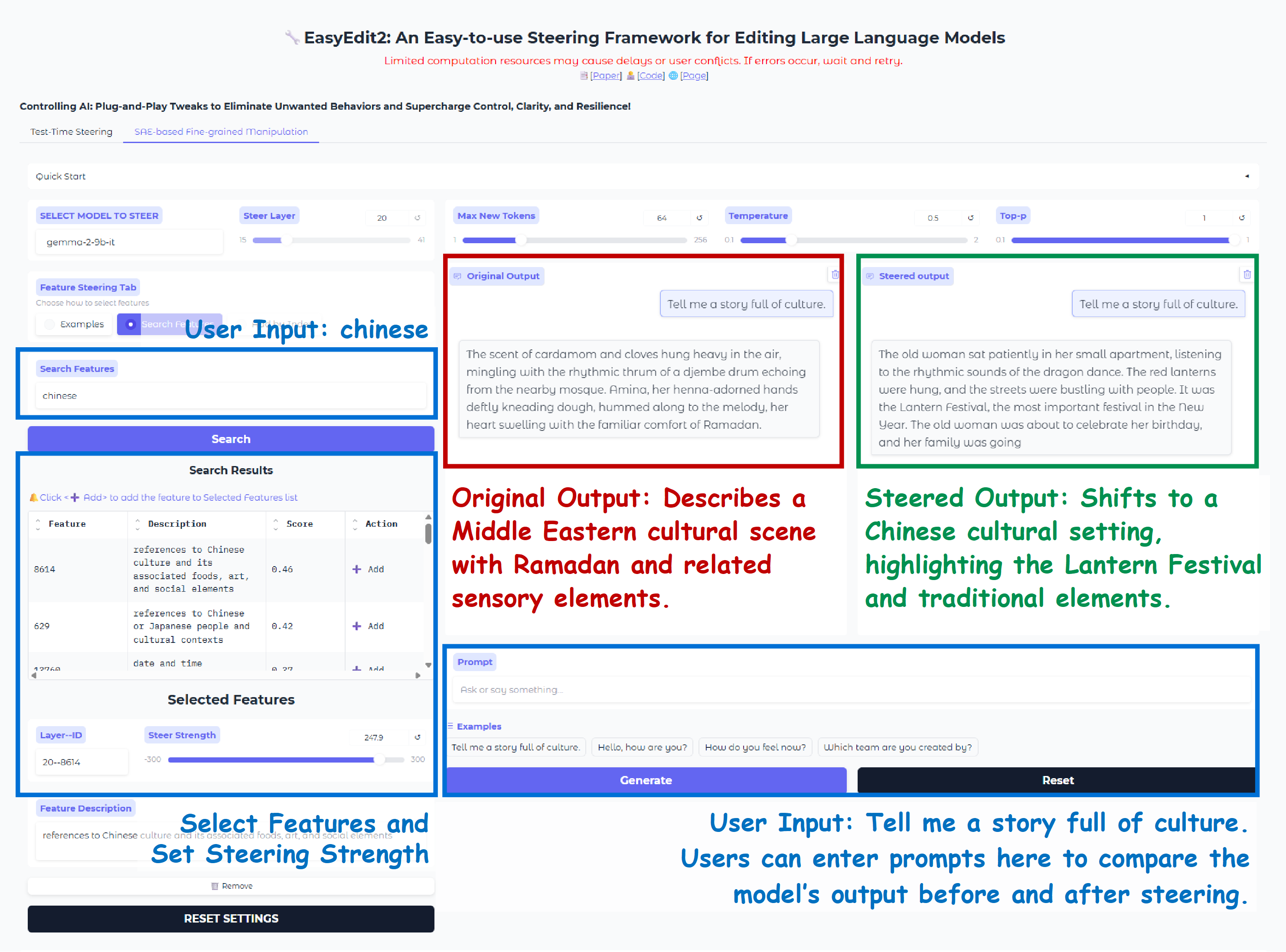}
    \caption{SAE-based fine-grained manipulation tab, showing feature-based steering for model control. 
    In this example, the user inputs ``chinese'' to search for related SAE features. 
    After selecting the relevant features, the steering strength is set, and the user provides the prompt ``Tell me a story full of culture.'' 
    The resulting output is steered to emphasize Chinese culture, including food, art, and social elements, demonstrating the effectiveness of fine-grained steering in guiding the model's response based on cultural context.}
    \label{fig:sae_feature_steering}
\end{figure*}

Figure~\ref{fig:sae_feature_steering} shows the SAE-based fine-grained control tab in online demo, where users search for features and adjust steering strength to modify outputs.

\end{document}